\pdfoutput=1

\documentclass[11pt]{article}

\usepackage{amsmath,amssymb,amsfonts}
\usepackage{algorithmic}
\usepackage{graphicx}
\usepackage{textcomp}
\usepackage{xcolor}
\usepackage{tabularx}
\usepackage{subcaption,multicol,multirow,makecell,booktabs}
\usepackage[hidelinks]{hyperref}
\usepackage[numbers]{natbib}

\newcolumntype{Y}{>{\centering\arraybackslash}X}

\newcommand{\br}[1]{\left( #1 \right)}

\newcommand{\set}[1]{\left\{ #1 \right\}}

\newcommand{\vfour}[4]{
\renewcommand*\env@matrix[1][\arraystretch]{%
  \edef\arraystretch{#1}%
  \hskip -\arraycolsep
  \let\@ifnextchar\new@ifnextchar
  \array{*\c@MaxMatrixCols c}}
\begin{bmatrix}
	#1 \\
	#2 \\
	#3 \\
	#4
\end{bmatrix}}
\makeatletter

\newcommand{\specialcell}[2][c]{%
\begin{tabular}[#1]{@{}c@{}}#2\end{tabular}} 

\newcolumntype{R}{>{\raggedleft\arraybackslash}X}
\newcolumntype{L}{>{\raggedright\arraybackslash}X}
\newcolumntype{s}{>{\hsize=.35\hsize}L}

\def\namedlabel#1#2{\begingroup
    #2%
    \def\@currentlabel{#2}%
    \phantomsection\label{#1}\endgroup
}

\def\BibTeX{{\rm B\kern-.05em{\sc i\kern-.025em b}\kern-.08em
    T\kern-.1667em\lower.7ex\hbox{E}\kern-.125emX}}

\usepackage[final]{acl}

\usepackage{times}
\usepackage{latexsym}

\usepackage[T1]{fontenc}

\usepackage[utf8]{inputenc}

\usepackage{microtype}

\usepackage{inconsolata}

\title{Personality Profiling: How informative are social media profiles in predicting personal information?}

\author{Joshua Watt, Lewis Mitchell and Jonathan Tuke \\
  \textit{School of Computer \& Mathematical Sciences, The University of Adelaide}, Adelaide SA 5005, Australia \\
  \texttt{\{joshua.watt,simon.tuke,lewis.mitchell\}@adelaide.edu.au}}

\begin{document}
\maketitle
\begin{abstract}

Personality profiling has been utilised by companies for targeted advertising, political campaigns and public health campaigns. 
However, the accuracy and versatility of such models remains relatively unknown. 
Here we explore the extent to which peoples' online digital footprints can be used to profile their Myers-Briggs personality type. 
We analyse and compare four models: logistic regression, naive Bayes, support vector machines (SVMs) and random forests. 
We discover that a SVM model achieves the best accuracy of 20.95\% for predicting a complete personality type. 
However, logistic regression models perform only marginally worse and are significantly faster to train and perform predictions. 
Moreover, we develop a statistical framework for assessing the importance of different sets of features in our models. 
We discover some features to be more informative than others in the Intuitive/Sensory ($p = 0.032$) and Thinking/Feeling ($p = 0.019$) models. 
Many labelled datasets present substantial class imbalances of personal characteristics on social media, including our own. 
We therefore highlight the need for attentive consideration when reporting model performance on such datasets and compare a number of methods to fix class-imbalance problems.

\end{abstract}

\section{Introduction}
In 2023 there are over 4.59 billion social media users worldwide, constituting approximately 60\% of the world's population \citep{dixonNumberWorldwideSocial2022}. 
This enables most of the world to be connected, creating an online \emph{information environment}. 
The huge amounts of individual-level data provided by each user is an important aspect of social media which is unique to this type of information environment. 
Consequently, it is crucial for scholars to understand how this aspect of social media may impact society. 
There exists a need to quantify the extent to which social media can be weaponized by governments and other organisations for influence.

Every time a user enters a social media application, they leave a unique data trace -- information they have posted, liked, shared, commented, even how long they have spent viewing different material on the application. 
We refer to this unique trace of data as a user's online digital footprint.
It has been suggested that someone's online digital footprint can expose actionable information about them, including their personality profile, relationship status, political opinions and even their propensity to adopt a particular opinion or behavior \citep{wylieMindfCkCambridge2020,patilPersonalityPredictionUsing2021,tadessePersonalityPredictionsBased2018,tanderaPersonalityPredictionSystem2017,weber2020arsonemergency,tuke2020pachinko}. 
Cambridge Analytica was suggested to use online digital footprints to impact the result of the 2016 US election and the 2016 Brexit referendum \citep{wylieMindfCkCambridge2020}. 
However, the extent to which companies like Cambridge Analytica can determine this information from social media data is still questioned \citep{patilPersonalityPredictionUsing2021,tadessePersonalityPredictionsBased2018,tanderaPersonalityPredictionSystem2017}. 
As a result, it is of interest for individuals to understand the extent of information that is attainable from their online digital footprint. 
This is also of key concern for governments, who seek to maintain democracies and the ethical use of such data.

We seek to determine how informative online digital footprints are in predicting Myers-Briggs personality types. 
This is a theoretical model comprised of four traits/dichotomies, based on Jungian theory \citep{blockHowMyersBriggsPersonality2018,jungCollectedWorksJung1976}. 
Modelling personal information about individuals using their online information has previously enabled researchers to understand the accuracy of such models. 
We extend this work by creating a new labelled dataset of Myers-Briggs personality types on Twitter and a statistical modelling framework which can be generally applied to any labelled characteristic of online accounts. 
We aim to reconsider the personality profiling and political microtargetting performed by companies like Cambridge Analytica.

First we collect a labelled dataset of accounts with self-reported Myers-Briggs personality types. 
We then collect a number of different features for these accounts including social metadata features and linguistic features: LIWC \citep{pennebakerDevelopmentPsychometricProperties2015}; VADER \citep{huttoVADERParsimoniousRulebased2015a}; BERT \citep{devlinBERTPretrainingDeep2019}; and Botometer \citep{sayyadiharikandehDetectionNovelSocial2020}. 
We then create independent logistic regression (LR), naive Bayes (NB), support vector machines (SVMs) and random forests (RF) models on each dichotomy to model the Myers-Briggs personality type of the accounts. 
As part of this, we consider four different weighting/sampling techniques to adjust for class imbalances. 
Lastly, we provide a statistical framework for analysing the importance of different features in these models. 
We consider the importance of features at an individual level and across groups of features for each dichotomy. 
Our main contributions are:
(i) A labelled dataset\footnote{Dataset available at \url{https://figshare.com/articles/dataset/Self-Reported_Myers-Briggs_Personality_Types_on_Twitter/23620554?file=41445756}.} of 68,958 Twitter users along with their Myers-Briggs personality types, the largest available dataset (to our knowledge) of labelled Myers-Briggs personality types on Twitter \citep{Watt2023};
(ii) A statistical framework to combine NLP tools and mathematical models to predict online users' personality types, which can be more broadly used to model any labelled characteristics about online accounts;
(iii) A comparison of machine learning models on NLP features, and a comparison of various weighting/sampling techniques to address problems with class imbalance;
(iv) Statistical methods which compare the importance of different features in NLP-based models at an individual level and across groups of features.

\section{Background}
Myers-Briggs \citep{blockHowMyersBriggsPersonality2018} is the most well-known personality model, being applied in hiring processes, social dynamics, education and relationships \citep{devriesMainDimensionsSport2020,walshTheoryPersonalityTypes1992,laneBlackwellHandbookGlobal2009}. 
The Myers-Briggs Type Indicator (MBTI) handbook illustrates a four factor model of personality where people form their `personality type' by attaining one attribute from each of four dichotomies; Extrovert/Introvert, Intuitive/Sensory, Thinking/Feeling and Judging/Perceiving. 
This gives 16 different personality types where a letter from each dichotomy is taken to produce a four letter acronym, e.g., `ENTJ' or `ISFP'.

The model has received substantial scrutiny, particularly from psychologists who question its validity and reliability \citep{pittengerMeasuringMBTIComing1993, grantGoodbyeMBTIFad}. 
Nonetheless, we utilise the Myers-Briggs model in our analysis for the following reasons:
(i) Thousands of Twitter users self-report their MBTI on Twitter. This enables us to obtain a labelled dataset through appropriately querying for each of the 4 letter personality type acronyms that are unique to MBTI.
(ii) The Myers-Briggs model has the largest number of self-reports on Twitter, enabling us to achieve the largest labelled personality dataset on Twitter.
(iii) We aim to develop a framework for modelling personality profiles from social media data using statistical machine learning (ML) approaches. MTBI is a test case for our framework, which can be applied to other personality models (or other labelings/characteristics of individuals on social media) more generally.

Open-source labelled training data with Myers-Briggs personality types has not existed until recently. \citet{plankPersonalityTraitsTwitter2015} modeled the MBTI of Twitter users through attaining a small dataset of 1,500 users and \citet{gjurkovicRedditGoldMine2018} modeled the MBTI on a larger corpus of Reddit users. 
In 2017, \citet{jollyMBTIMyersBriggsPersonality} posted a labelled MBTI dataset on Kaggle, constituting the only known publicly available labelled dataset used for modelling the MBTI of social media users. 
The dataset was comprised of 8,675 users, their personality types and a section of their last 50 posts on an online forum called \url{personalitycafe.com}. 
This small online forum contains 153,000 members dedicated to discussing health, behavior, personality types and personality testing. 
The discussions are therefore quite different to those on other social media platforms, and likely a different demographic. 
Hence, this dataset is likely not generalisable to other platforms like Twitter and Facebook. 
It is also relatively small and imbalanced, limiting which models can be utilised on various feature sets.
Class imbalance is considerable in all cases, and in one particular dataset some classes are up to 28 times larger than their counterpart. 
Nevertheless, many papers apply machine learning models to such datasets without accounting for these class imbalances \citep{tadessePersonalityPredictionsBased2018, basaranNeuralNetworkApproach2021, kehMyersBriggsPersonalityClassification2019, amirhosseiniMachineLearningApproach2020, patilPersonalityPredictionUsing2021}.
Consequently, the metrics reported often misrepresent model performance, and instead highlight the severity of class imbalances in the datasets. 

\section{Data Collection \& Preprocessing}\label{s:data_collection}
We discovered a number of Twitter accounts self-report their MBTI on Twitter as a regular expression. We therefore formulated two methods for querying and labelling the Myers-Briggs personality type of accounts. 
Let $\Omega$ define the set of 16 acronyms for Myers-Briggs personality types.
\begin{enumerate}
    \item[\namedlabel{item:data_collection_M1}{\textbf{M1}}] Query: ${\tt \set{x : x \in \Omega}}$. We obtained the set of users who currently self-report their personality type in their username or biography.
    \item[\namedlabel{item:data_collection_M2}{\textbf{M2}}] $\text{Query: } {\tt \{\br{\text{I am } x} \vee \br{\text{I am a } x} \vee \br{\text{I am an } x}}$ $\tt{: x \in \Omega\}}$. We obtained the set of users who have self-reported their personality type in a Tweet since Twitter's creation (March 26, 2006). Note that we only searched for self-reports in Tweets, not Retweets, Quotes and Replies -- due to a number of users often not self-reporting their own MBTI when referencing MBTI acronyms in these forms of communication.
\end{enumerate}
Queries were not case-sensitive.

The resulting labelled dataset comprised of 68,958 users; the dataset and more details on its collection are provided in \citep{Watt2023}.
We collected 15,986 accounts by querying usernames and biographies, and 52,972 accounts from querying tweets, with misclassification rates 1.9\% and 3.4\% based on random samples of 1,000 accounts from each.


Next we obtained account characteristics for each user, including their biography, most recent 100 tweets/quotes, as well as a set of Social Metadata (SM) features. 
The user's biography and the 100 tweets/quotes were used to generate a set of linguistic features, whereas SM features (Table \ref{tab:features}) are directly used as numeric features in the models. 

We removed duplicate users, then combined the biography and tweets into a combined text for every account. 
We then:
1. Normalised the text and calculated each account's dominant language.
2. Removed non-English language using the Compact Language Detect 2 (PyCLD2) library.
3. Calculated (language-dependent) Botometer scores\footnote{Further discussion: \url{https://rapidapi.com/OSoMe/api/botometer-pro/details}}.
4. Converted text to lowercase, removed URLs, email addresses, punctuation and numbers.
5. Tokenized using the Tweet Tokenizer from the Natural Language Toolkit (NLTK) \citep{birdNaturalLanguageProcessing2009}.
6. Removed empty tokens and any instances of the 16 MBTI acronyms.

Next, we formulated an inclusion-exclusion criteria to determine whether a personality could be profiled from a Twitter account -- we kept accounts with over 100 tweets/quotes, over 50\% English language, Botometer CAP score less than 0.8, and strictly one MBTI type referenced.


We use the Botometer CAP score because we are interested in the overall bot likelihood and not the sub-category bot likelihoods. 
Unfortunately, there is no consistency in the literature on thresholds for binary bot classification. 
Rather, authors define their threshold based on a false positive rate in the context of their problem. 
For instance, \citet{wojcikBotsTwittersphere2018} use a threshold of $0.43$ for their political analysis of the twittersphere, whereas \citet{kellerSocialBotsElection2019} use a larger threshold of $0.76$ for their analysis of social bots in election campaigns. 
To avoid large numbers of false positive bot classifications, we chose a high threshold of $0.8$.

Finally, we extracted the LIWC, BERT and VADER features from the text. 
The data cleaning techniques above were performed only for LIWC feature extraction, whereas the BERT and VADER features can be extracted directly from the raw text output. 
Thus, we calculated the LIWC features on the combined text by micro-averaging the tokens present in each LIWC category for every user. 
Next, we calculated the BERT features on the raw Twitter output using BERTweet \citep{nguyenBERTweetPreTrainedLanguage2020}, a pre-trained language model for English Tweets. 
First, we averaged the embeddings for the tokens to form a single embedding vector for each tweet/quote, then averaged the embedding vectors for the tweets/quotes to create a single 768-dimensional embedding vector for each user. 
We calculated the VADER features (sentiment, proportion of positive words and proportion of negative words) on the raw Twitter output for each user and include scores for both a user's biography and their tweets. 
We distinguish these because of contextual differences in the language; biographies often discuss oneself and tweets often discuss one's environment. 
We then have a total of 866 features; these are provided in Table \ref{tab:features}.

\begin{table}[ht!]
\scriptsize
    \begin{center}
        \begin{tabularx}{\linewidth}{ l|R } 
            Category & Features \\
            \hline
            SM & followers\_count, friends\_count, listed\_count, favourites\_count, geo\_enabled, verified, statuses\_count, default\_profile, default\_profile\_image, profile\_use\_background\_image, has\_extended\_profile \\
            \hline
            Botometer & cap\_english, english\_astroturf, english\_fake\_follower, english\_financial, english\_other, english\_self\_declared, english\_spammer \\
            \hline
            LIWC &  function, pronoun, ppron, i, we, you, shehe, they, ipron, article, prep, auxverb, adverb, conj, negate, verb, adj, compare, interrog, number, quant, affect, posemo, negemo, anx, anger, sad, social, family, friend, female, male, cogproc, insight, cause, discrep, tentat, certain, differ, percept, see, hear, feel, bio, body, health, sexual, ingest, drives, affiliation, achiev, power, reward, risk, focuspast, focuspresent, focusfuture, relativ, motion, space, time, work, leisure, home, money, relig, death, informal, swear, netspeak, assent, nonflu, filler, total\_word\_count \\
            \hline
            BERT & $\set{e_i \; ; \; i = 1,\dots,768}$ \\
            \hline
            VADER & tweets\_sentiment, bio\_sentiment, tweets\_pos\_words, bio\_pos\_words, tweets\_neg\_words, bio\_neg\_words \\
        \end{tabularx}
    \end{center}
    \caption{Features in our models, separated by category.}
    \label{tab:features}
\end{table}

\section{Exploratory Data Analysis}
\begin{figure}[ht]
  \begin{center}
    \includegraphics[width=\linewidth]{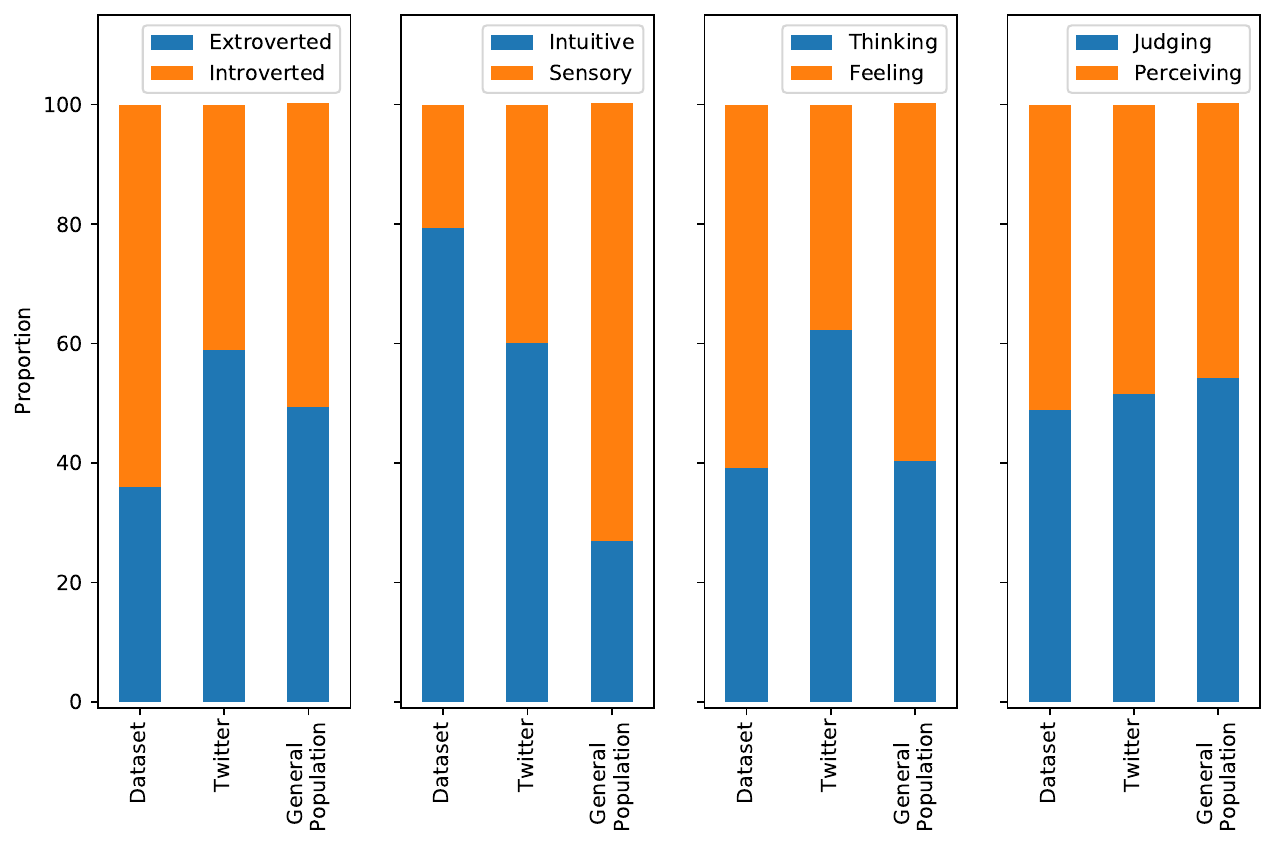}
  \end{center}
  \caption{Proportion of accounts displaying each dichotomous trait in our dataset, on Twitter and in the general population.}
  \label{fig:class_proportions}
\end{figure}

We performed an exploratory data analysis (EDA) on the dataset to determine important information about our dataset, prior to any modelling. 
We acknowledge and discuss two forms of potential bias in our dataset: (i) only considering MBTI types on Twitter; (ii) only selecting accounts which satisfy our inclusion-exclusion criteria as well as self-report their MBTI types on Twitter.
Figure \ref{fig:class_proportions} demonstrates these biases through bar plots showcasing the proportions of the MBTI dichotomies in our dataset. 
We compare with a study reporting MBTI proportions on Twitter \citep{schaubhutMyersBriggsTypeSocial2012}, and with the proportion of personality types in the general population \citep{robinsonHowRareYour1998}.

A noticeable imbalance in the Intuitive/Sensory dichotomy exists across all datasets in Figure \ref{fig:class_proportions}. 
There are also observable imbalances in the Extrovert/Introvert and Thinking/Feeling dichotomies, whereas the Judging/Perceiving dichotomy is more balanced across each dataset than the other dichotomies. 
The imbalances in our dataset are mostly consistent with those from \url{www.personalitycafe.com}. 
The higher proportion of introverts in our dataset is consistent with \citep{HowTechnologySocial2015} who find that introverts tend to use social media as a primary form of communication, whereas extroverts tend to prefer communicating in-person. 
The larger proportion of intuitives in our dataset is consistent with \citet{schaubhutMyersBriggsTypeSocial2012} who discovered that more Intuitive individuals (13\%) reported being active users of Twitter than individuals with a preference for Sensing (8\%). 
The imbalance in the Thinking/Feeling dichotomy in our dataset is opposite to what we observe in the Twitter dataset. 
However, \citet{schaubhutMyersBriggsTypeSocial2012} found that people displaying the Feeling trait are more likely to spend their personal time browsing, interacting and sharing information on Facebook. 
Provided the same is true for Twitter users, our inclusion-exclusion condition requiring users to be active on Twitter (i.e. tweet/quote at least 100 times) may bias our dataset leading to more users exerting the Feelings trait.

Some authors don't assume independence between the dichotomies \citep{basaranNeuralNetworkApproach2021,patilPersonalityPredictionUsing2021}, whereas most choose to model the dichotomies independently \citep{alamPersonalityTraitsRecognition2013,sumpterOutnumberedFacebookGoogle2018,bharadwajPersonaTraitsIdentification2018, kehMyersBriggsPersonalityClassification2019, amirhosseiniMachineLearningApproach2020}. 
We take a data-driven approach, determining the dependency structure of the four MBTI dichotomies in our dataset using the bias-corrected version of the Cramér's V Statistic \citep{alma991307001811} (Table \ref{tab:mbti_cramers_v}). 
The Cramér's V statistic is small in every case, implying that the four Myers-Briggs dichotomies are independent in our dataset, and so we model them independently.

\begin{table}[ht]
    \centering
    \begin{tabular}{c|cccc}
        & E/I & N/S & T/F & J/P \\
        \hline
        E/I & 1.00 & 0.03 & 0.00 & 0.10 \\
        N/S & 0.03 & 1.00 & 0.02 & 0.08 \\
        T/F & 0.00 & 0.02 & 1.00 & 0.11 \\
        J/P & 0.10 & 0.08 & 0.11 & 1.00
    \end{tabular}
    \caption{Pairwise results of the bias-corrected Cramér's V Statistic between the MBTI dichotomies for our dataset.}
    \label{tab:mbti_cramers_v}
\end{table}

We performed a Principal Component Analysis (PCA) on the features to discover if we could significantly reduce the dimension of the feature space, and multicollinearity between the features. 
The first principal component explains 25.1\% of the variance in the data and the first 200 principal components explain 95.4\% of the variance in the data. 
As a result, we utilise the first 200 PCA components in our machine learning models, significantly reducing both the dimension of the feature space and the multicollinearity of the features.


\section{Model Comparison}
We train LR, NB, SVM and RF classifiers on each of the four dichotomies in our dataset, using 10-fold cross validation. 
The class imbalances we observe for some dichotomies (particularly Intuitive/Sensory and Extrovert/Introvert), leads us to perform four different weighting/sampling techniques on the training data prior to model fitting:
(i) Weight the importance of classifying dichotomies,
(ii) Upsample the minority class (with replacement),
(iii) Perform the Synthetic Minority Oversampling Technique (SMOTE) on the minority class,
(iv) Downsample the majority class.

\begin{figure}[htb!]
        \centering
        \begin{subfigure}{\linewidth}
          \centering
          \includegraphics[width=.8\linewidth]{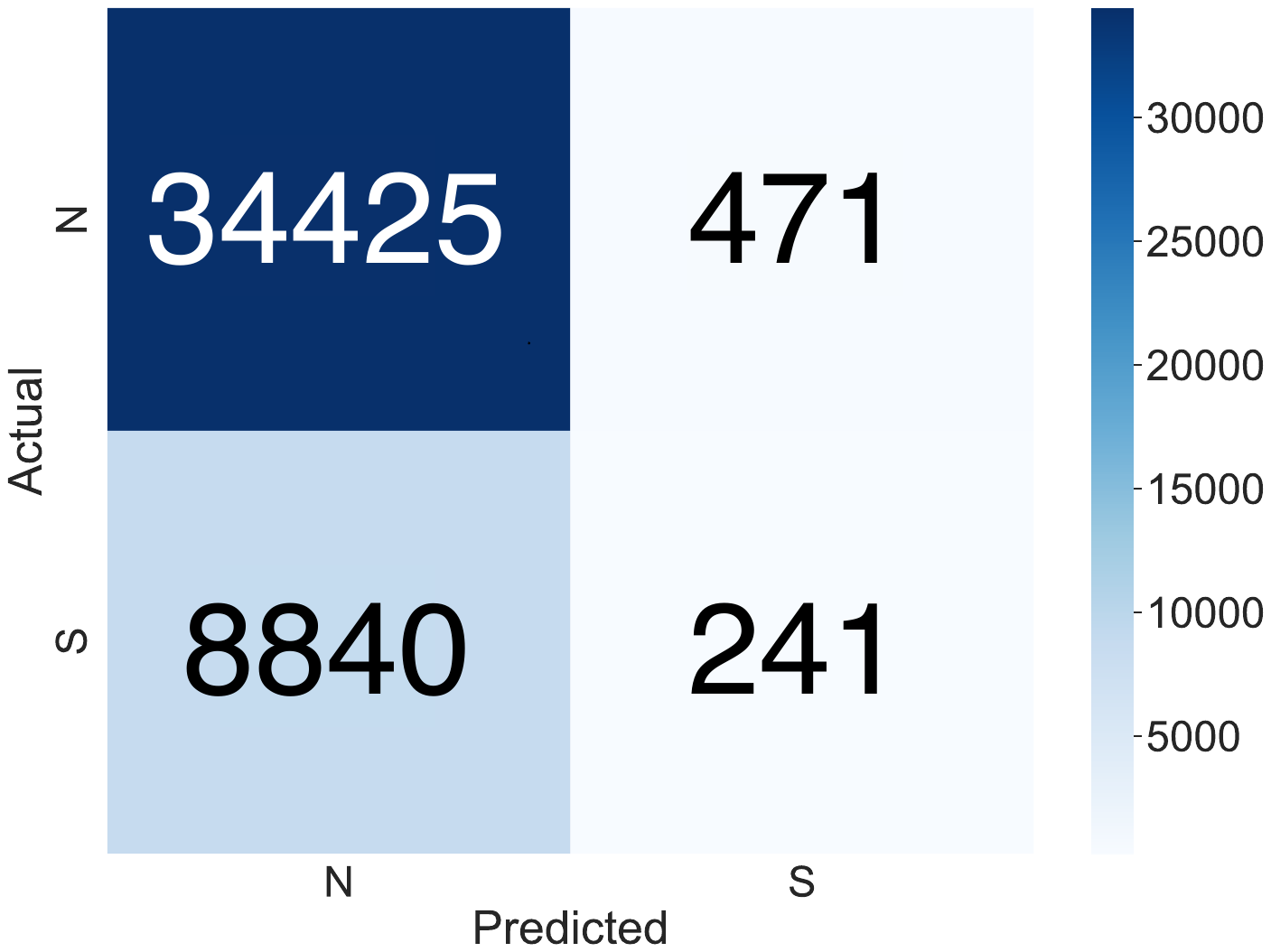}
          \caption{Standard logistic regression}
          \label{fig:log_reg_mat}
        \end{subfigure}
        \begin{subfigure}{\linewidth}
          \centering
          \includegraphics[width=.8\linewidth]{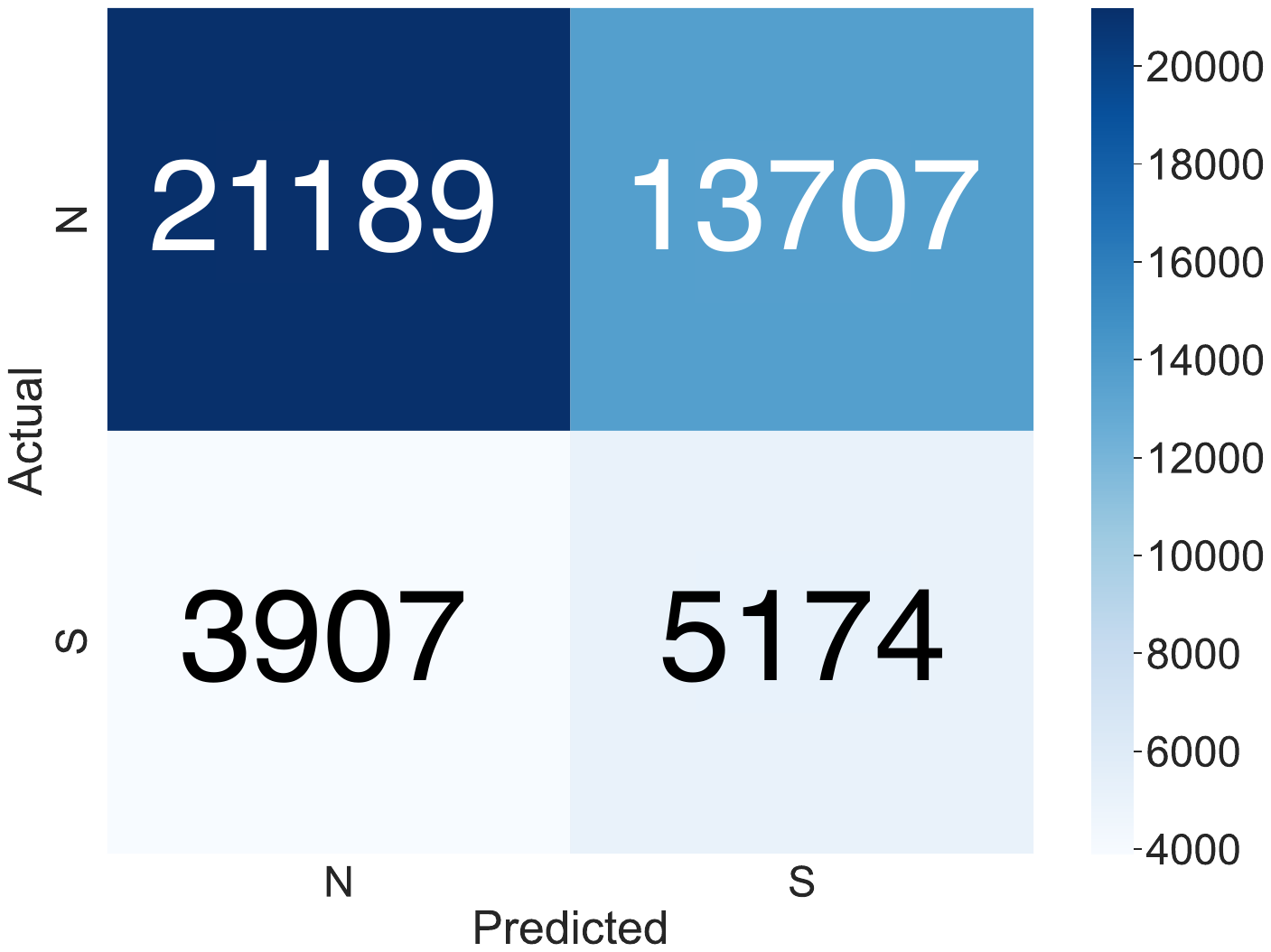}
          \caption{Upsampled logistic regression}
          \label{fig:upsampled_log_reg_mat}
        \end{subfigure}
        \caption{Confusion matrices for modelling the N/S dichotomy.}
        \label{fig:NS_conf_matrices}
\end{figure}

Each model uses the first 200 principal components of the features in Table \ref{tab:features} as predictors. 
As an example, Figure \ref{fig:NS_conf_matrices} shows confusion matrices for the Intuitive/Sensory dichotomy under the standard LR model and the upsampled LR model.

This shows that the standard LR model primarily predicts the majority class, indicating that it exploits the class imbalance to make predictions on the test sets. 
In comparison, the upsampled model predicts significantly more of the minority class on the test sets, resulting in more accurate predictions for the minority class. 
We observe similar behavior for all other models, highlighting the importance of weighting/sampling techniques to ameliorate the effect of class imbalance for prediction. 
However, we observe a clear trade-off between accurately predicting the majority and minority classes, with an overall reduction in accuracy due to weighting/sampling techniques. 
We therefore report both accuracy and Area Under the Curve (AUC) metrics for each of our models in Table \ref{tab:combined_accuracies}.  
We report four types of accuracy depending on the number of accurately predicted dichotomies in each model. 
Of course, accuracy can be a misleading metric when assessing a model's performance on unbalanced data, so for comparison we report the accuracies for a random classifier and a majority class classifier. 
Moreover, we use an approach similar to other authors to report two types of AUC for each model \citep{guncarApplicationMachineLearning2018, deUsingMachineLearning2020}: we macro-average and micro-average the true positive rate and false positive rate at each threshold of the ROC curve for the independent models of each dichotomy. 
This provides us with two ROC curves (and AUC metrics) for each model. 
The micro-averaged AUC aggregates the contributions of all samples in each model and weights individual predictions equally, so it is generally less sensitive to class imbalances. 
Table \ref{tab:combined_accuracies} compares the accuracies and AUCs of the best performing models from each method. 
In each case, we include the `Standard' model and the weighted/sampling model which achieves the highest sum of micro- and macro-averaged AUC.

\begin{table}[ht]
\scriptsize
  \begin{center}
      \begin{tabularx}{\linewidth}{l c * {5}{Y}}
       & \multicolumn{4}{c}{Accurately Predicted Dichotomies} & \multicolumn{2}{c}{AUCs}\\
        \cmidrule(l){2-7}
        Model & 4 & $\geq 3$ & $\geq 2$ & $\geq 1$ & Macro & Micro   \\ 
        \midrule
        Standard LR & 20.82 & \textbf{60.43} & 89.35 & 98.82 & 0.6688 & 0.6547 \\
        SMOTE LR & 13.89 & 48.63 & 82.51 & 97.65 & 0.6642 & \textbf{0.6620} \\
        \hline
        Standard NB & 14.20 & 49.17 & 81.91 & 97.40 & 0.5784 & 0.5867 \\
        Upsampled NB & 13.75 & 48.06 & 80.82 & 97.18 & 0.5861 & 0.5917 \\
        \hline
        Standard SVM & \textbf{20.95} & 60.25 & \textbf{89.64} & \textbf{98.90} & \textbf{0.6693} & 0.6518 \\
        SMOTE SVM & 13.56 & 48.61 & 82.54 & 97.61 & 0.6660 & 0.6554 \\
        \hline
        Standard RF & 19.69 & 57.96 & 88.69 & 98.67 & 0.6223 & 0.6273 \\
        Upsampled RF & 19.70 & 58.16 & 88.48 & 98.76 & 0.6305 & 0.6264 \\
        \hline
        Random Classifier & 6.250 & 31.25 & 68.75 & 93.75 & 0.5000 & 0.5000 \\
        Majority Class & 15.31 & 54.54 & 87.20 & 98.28 & 0.5000 & 0.5000 \\
        \end{tabularx}
\end{center}
\caption{Accuracies and AUCs for best performing models. 
We include the `Standard' model (with no weighting/sampling) and best performing weighted/sampling model. 
The `best performing weighted/sampling model' is based on the sum of macro- and micro-averaged AUC.}
\label{tab:combined_accuracies}
\end{table}

Table \ref{tab:combined_accuracies} highlights the relatively small improvement in accuracy achieved by each model in comparison to the majority class classifier. 
It is clear that our standard SVM model is the best performing model on average. 
However, this model is only 5.64\% more accurate at predicting a user's complete personality type compared to the majority class classifier. 
This is a reasonable and statistically significant improvement, but we remark based on the above discussion that the standard models are simply exploiting the class imbalances in our dataset. 
Moreover, we achieve similar accuracies to  \citet{plankPersonalityTraitsTwitter2015}, who produced the only other Twitter dataset of labelled MBTI's (to our knowledge). 
In particular, we achieve better accuracies for the T/F and J/P dichotomies, and only marginally worse accuracies for E/I and N/S -- further evidencing that our models perform similarly to others in the literature.


Interestingly, the standard LR model most accurately predicts at least three of four user dichotomies and is only marginally worse than SVM for all other metrics. 
The LR model is also significantly faster to train than the SVMs -- making it the model of choice on larger datasets.

The AUC is important in discussions of model performance, especially for unbalanced datasets. 
This is because it equally weights the TPR and FPR, making it more robust for unbalanced datasets compared to accuracy. 
Most of our AUCs lie around 0.65, apart from the NB Classifiers. In particular, the best performance for the macro-averaged and micro-averaged AUCs is the standard SVM and SMOTE LR model, respectively. 
These AUCs are significantly larger than for both the random and majority class classifiers, indicating a clear `signal' in our features. 
We therefore perform an in-depth analysis of feature importance next.

\section{Feature Importance}
We perform independent upsampled LR models on each of the four MBTI dichotomies because they performed well on our dataset (macro- and micro-averaged AUCs: 0.6676 and 0.6536). 
We choose an LR model because it is fast to train, and straightforward to interpret and perform feature selection on. 
Moreover, we use an upsampled model because it does not involve creating `synthetic' data in the same way that SMOTE does -- this is important for determining feature importance.

\begin{figure}[h!]
    \centering
    \begin{subfigure}{0.9\linewidth}
      \centering
      \includegraphics[width=0.99\linewidth]{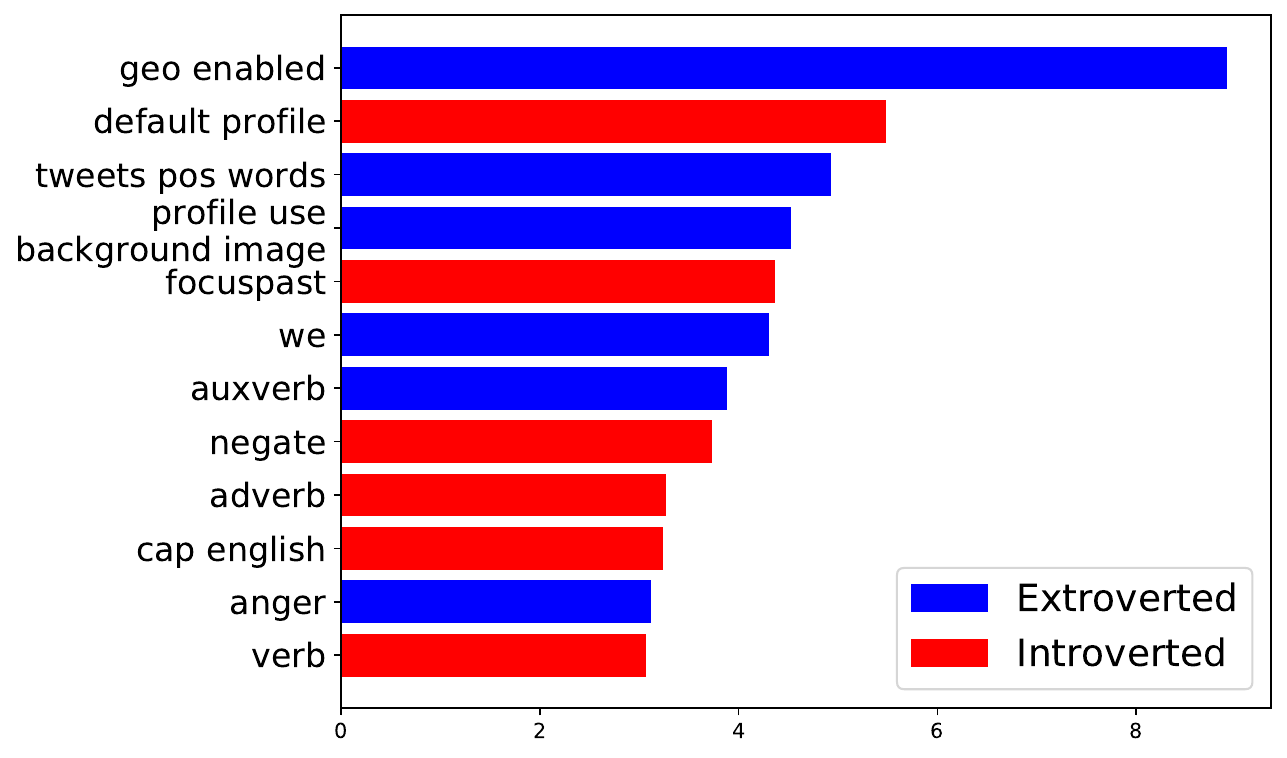}
      \caption{Extroverted/Introverted}
      \label{fig:EI_vip}
    \end{subfigure}
    \begin{subfigure}{0.9\linewidth}
        \centering
        \includegraphics[width=0.99\linewidth]{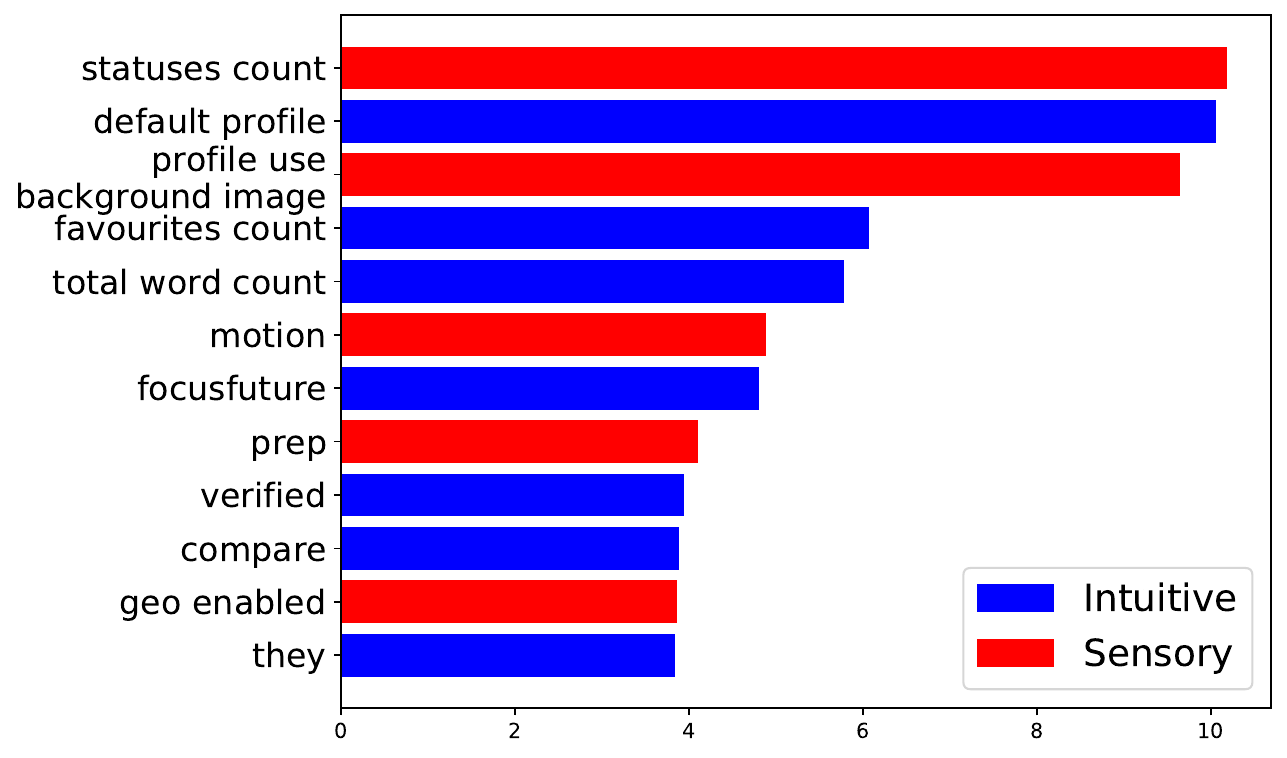}
        \caption{Intuitive/Sensory}
        \label{fig:NS_VIP}
      \end{subfigure}
    \begin{subfigure}{0.9\linewidth}
        \centering
        \includegraphics[width=0.99\linewidth]{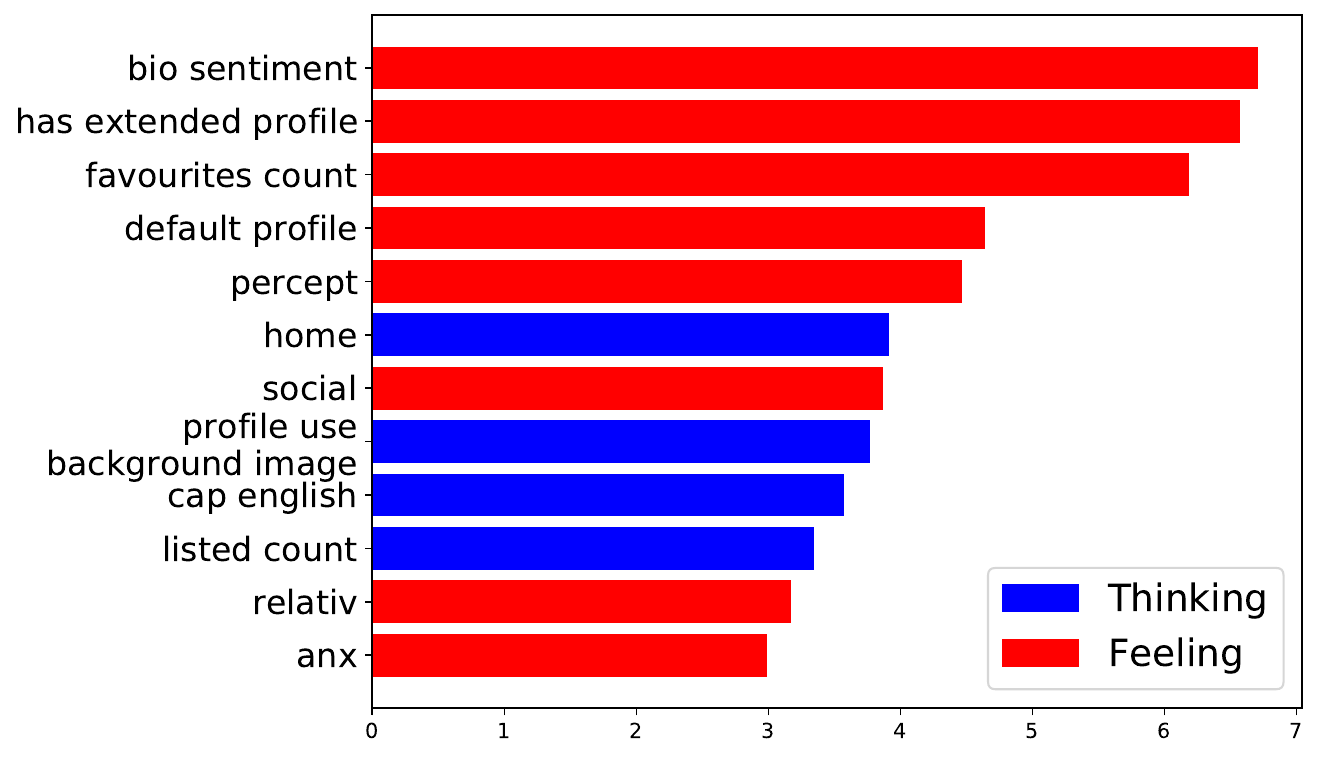}
        \caption{Thinking/Feeling}
        \label{fig:TF_VIP}
      \end{subfigure}
    \begin{subfigure}{0.9\linewidth}
      \centering
      \includegraphics[width=0.99\linewidth]{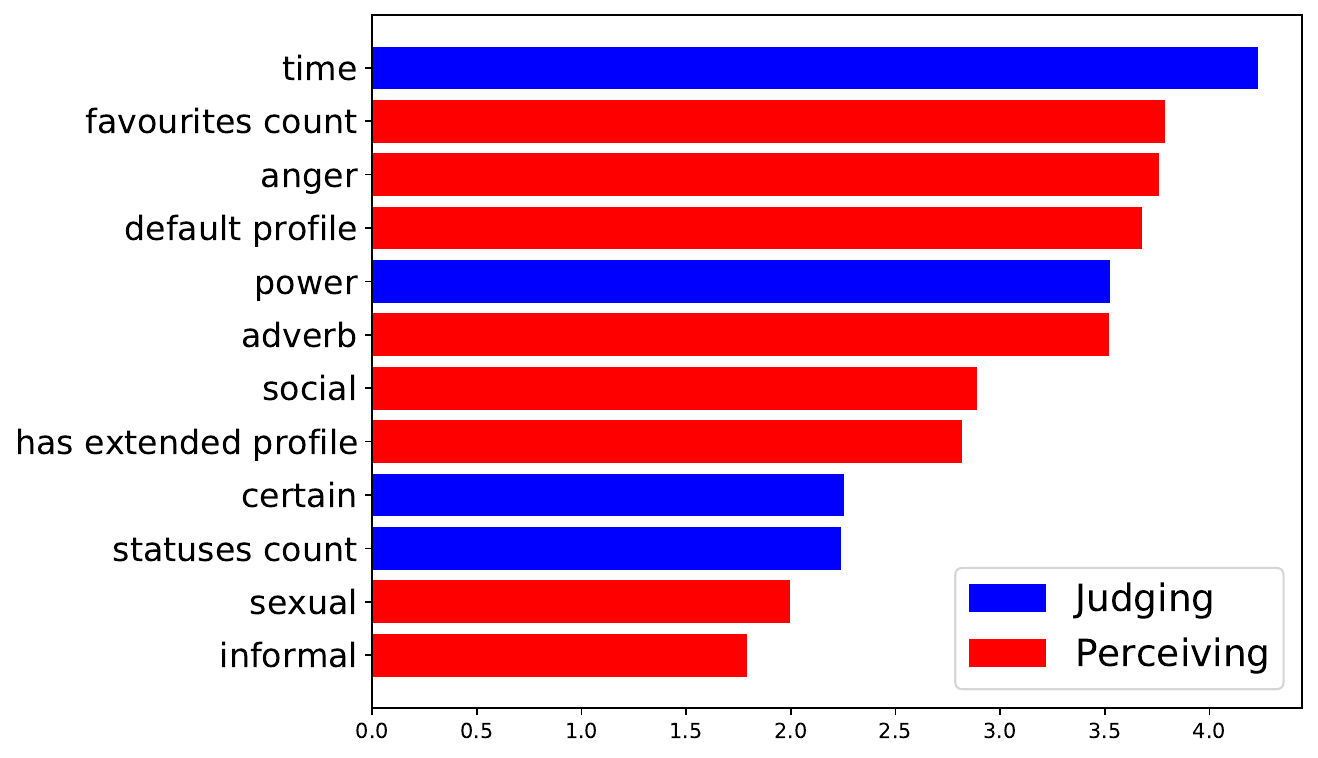}
      \caption{Judging/Perceiving}
      \label{fig:JP_VIP}
    \end{subfigure}
    \caption{Variable Importance Plots for an upsampled LR model for each dichotomy. Variables sorted by the absolute value of variable importance. Bars coloured by feature preference for each class.}
    \label{fig:VIP_plots}
\end{figure}

We consider the variable importance of the descriptive features in our models; these include all features except from BERT. 
For each dichotomy we fit the upsampled LR model and perform a stepwise feature selection to obtain a model with only significant features. 
In each case, we start with a null model and perform the stepwise selection algorithm on the $p$-values with a threshold in of $0.05$ and a threshold out of $0.1$. 
We determine the variable importance of features using the $t$-statistic for the parameter coefficients associated with each feature. 
For each dichotomy, we calculate the variable importance of each remaining feature after stepwise selection is complete, and display the absolute value of variable importance. 
Figure \ref{fig:VIP_plots} displays the 12 most important features for each model. 
We colour the bars based on the variable's preference for each class in the dichotomy.

\citet{pennebakerCognitiveEmotionalLanguage1996} suggested function words such as pronouns (pronoun), personal pronouns (ppron), 1st person singular (i), 1st person plural (we), prepositions (prep), auxiliary verbs (auxverb) and negations (negate), can describe people. 
Figure \ref{fig:VIP_plots} shows the function words that are significant predictors in our models, e.g., 1st person plurals are significant in the E/I model and prepositions are significant in the N/S model. 
This reinforces the importance of function words, and that techniques such as stop-word removal may remove useful information, particularly for tasks like personality prediction.

Extroverts tend to be associated with more positive language, and introverts have more focus on the past. 
Similarly, \citet{chenMetaanalysisLinguisticMarkers2020} suggested that extroverts display more positive emotion because they have a ``dispositional tendency to experience positive emotions''. 
Accounts with larger favourites count (i.e. the account likes more tweets) tend to be more intuitive, whereas accounts which write more statuses tend to be more sensory. 
Interpreting favourites as a proxy for the amount of information an account consumes, our results suggest that intuitives consume more information on Twitter, whereas sensory individuals write more. 
This proxy is of course not perfect, because people may consume information without liking it. 
Nonetheless, it is consistent with Myers-Briggs Foundation definitions, which state that intuitives pay ``most attention to impressions or the meaning and patterns of the information'', whereas sensors pay ``attention to physical reality, what I see, hear, touch, taste, and smell'' \citep{MyersBriggsFoundation2022}. 
The strongest predictor for the J/P dichotomy (Figure \ref{fig:JP_VIP}) is time; judgers are more likely to use words related to time and certainty compared to perceivers. 
`End', `until' and `season' are examples of time-related words and `always', `never' are words related to certainty. 
This is consistent with the Myers-Briggs Foundation, which states judgers ``prefer a planned or orderly way of life, like to have things settled and organized'' \citep{MyersBriggsFoundation2022}.

\begin{figure}[h!]
  \centering
  \begin{subfigure}{.7\linewidth}
    \centering
    \includegraphics[width=0.99\linewidth]{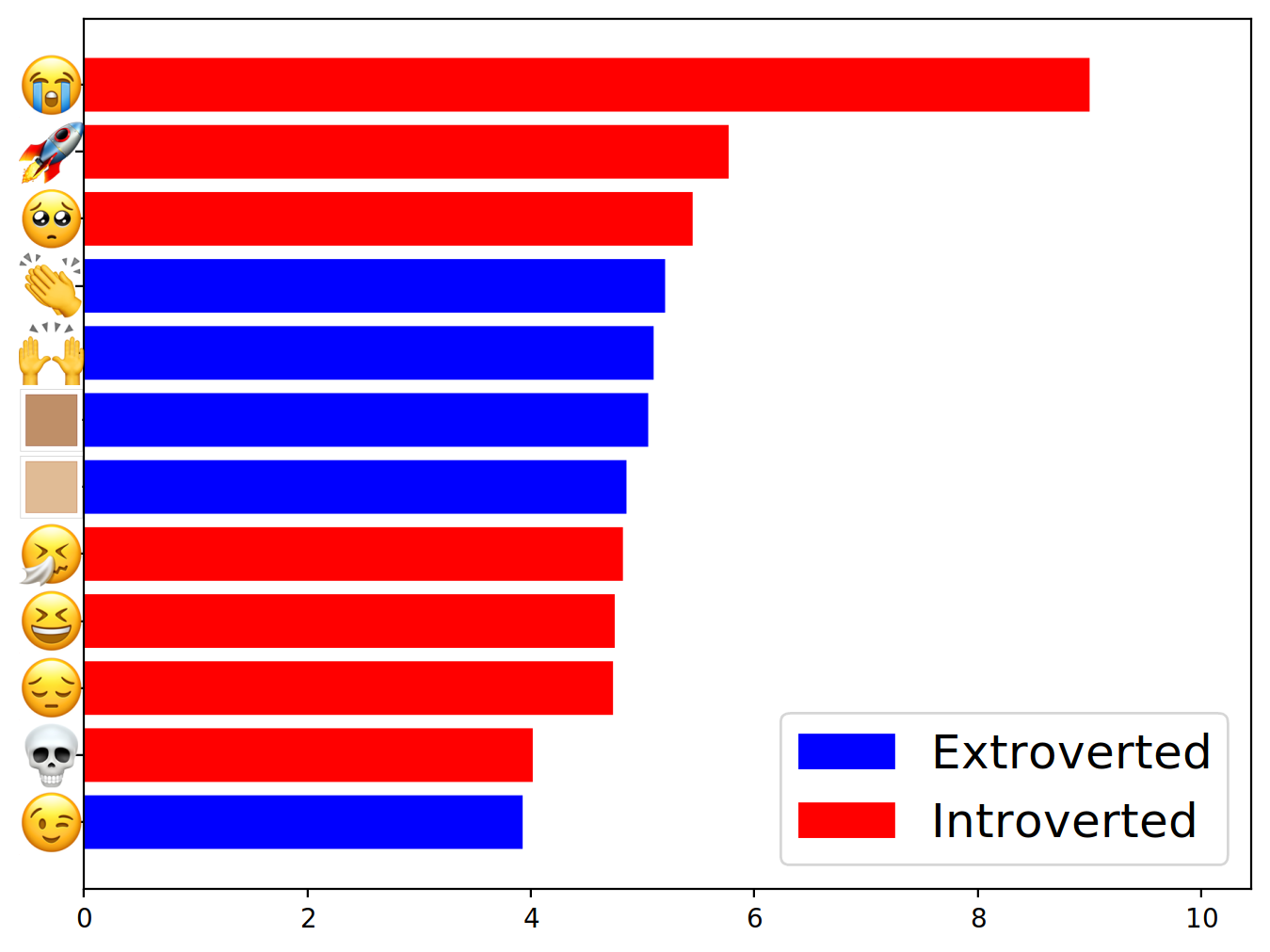}
    \caption{Extroverted/Introverted}
    \label{fig:EI_VIP_emoji}
  \end{subfigure}
  \begin{subfigure}{.7\linewidth}
      \centering
      \includegraphics[width=0.99\linewidth]{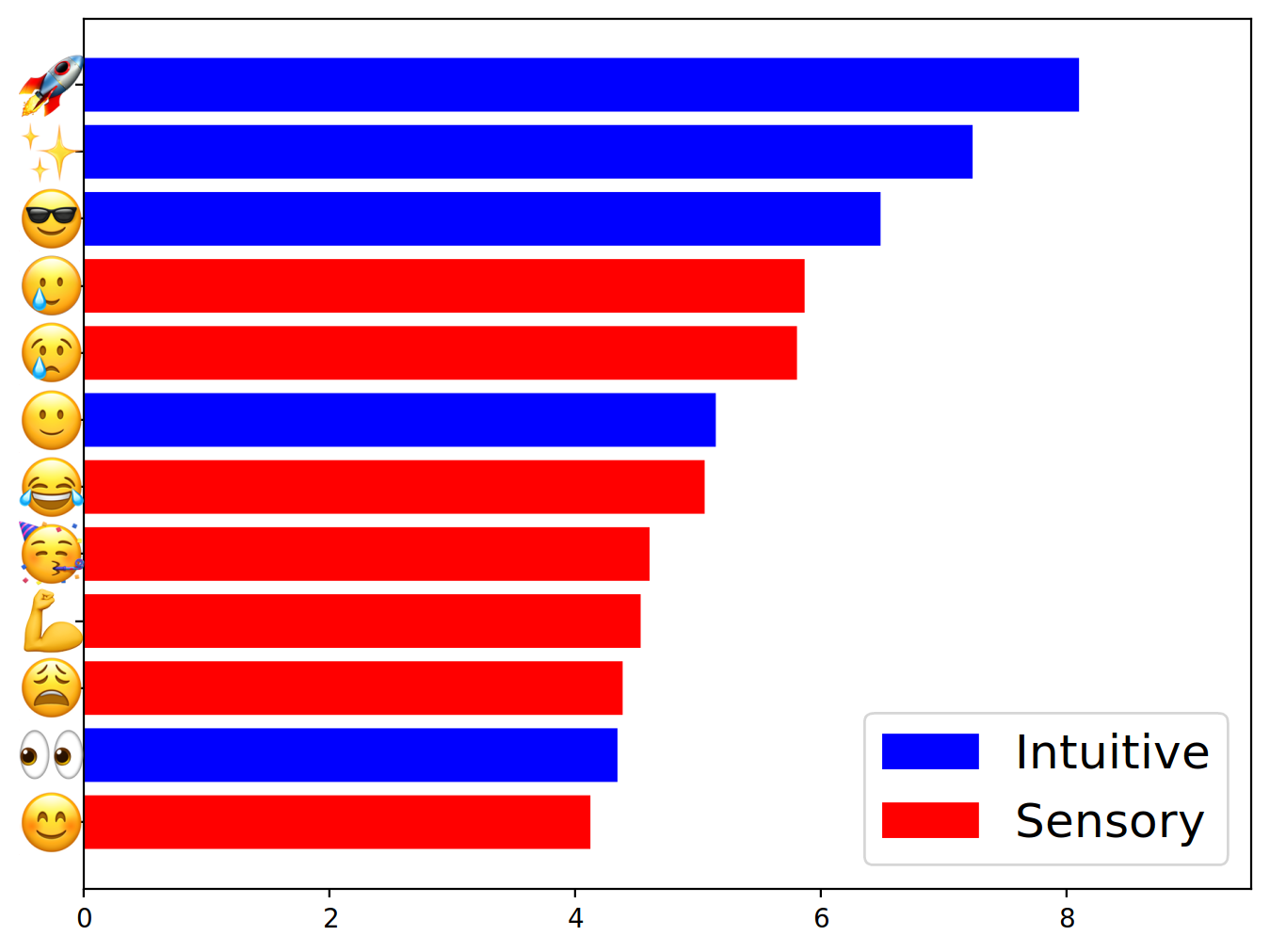}
      \caption{Intuitive/Sensory}
      \label{fig:NS_VIP_emoji}
    \end{subfigure}
  \begin{subfigure}{.7\linewidth}
      \centering
      \includegraphics[width=0.99\linewidth]{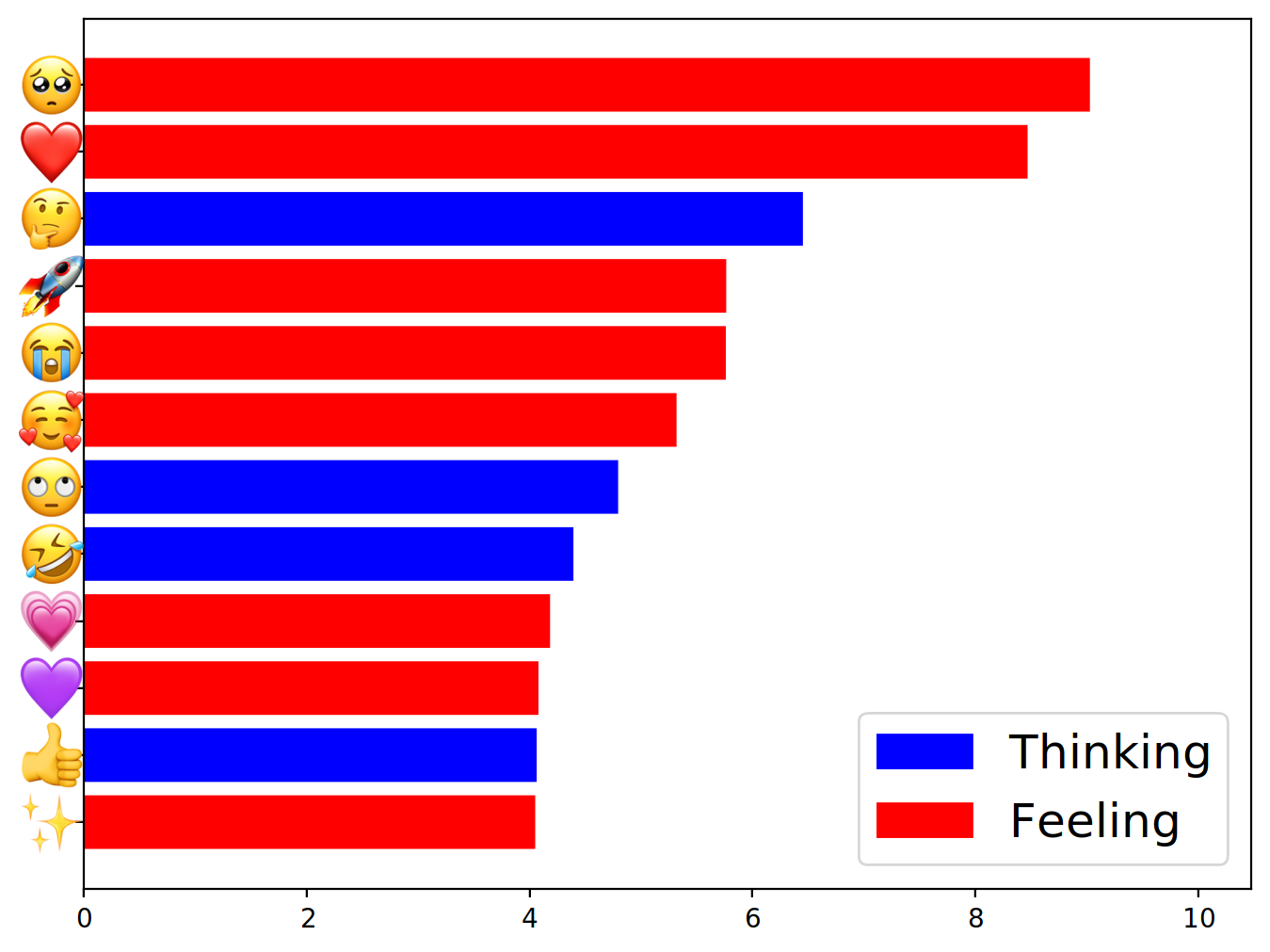}
      \caption{Thinking/Feeling}
      \label{fig:TF_VIP_emoji}
    \end{subfigure}
  \begin{subfigure}{.7\linewidth}
    \centering
    \includegraphics[width=0.99\linewidth]{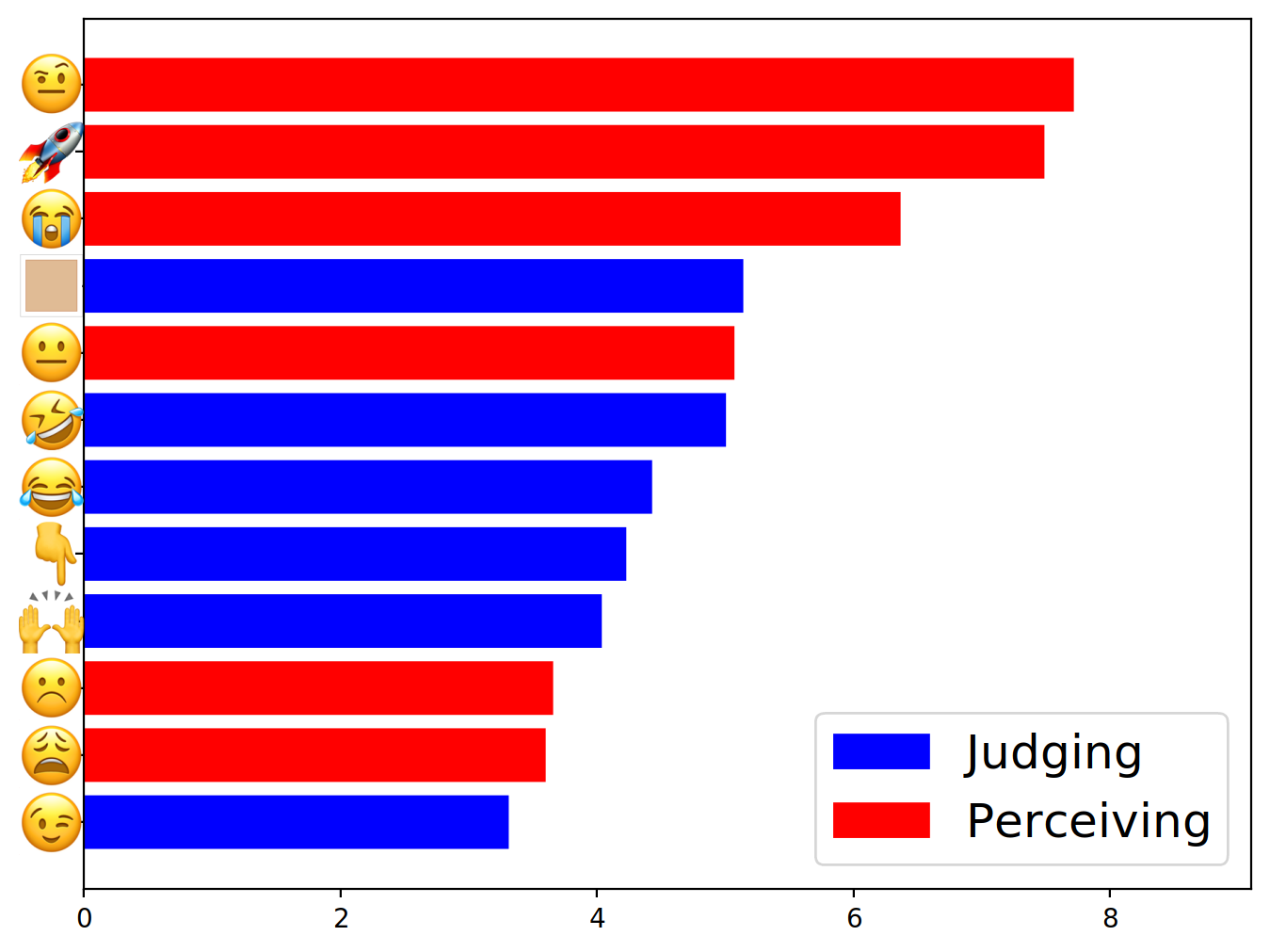}
    \caption{Judging/Perceiving}
    \label{fig:JP_VIP_emoji}
  \end{subfigure}
  \caption{Variable Importance Plots for emoji counts in the upsampled LR models. Variables sorted by absolute value of variable importance. We colour bars by the feature preference for each class.}
  \label{fig:VIP_plots_emoji}
\end{figure}

Next we explore how emoji usage relates to a Twitter user's MBTI. 
On Twitter, emojis often have multiple meanings. 
For instance, the rainbow flag can indicate support for  LGBTQ+ social movements, the wave can symbolise a “Resister” crowd of anti-Trump Twitter, and the okay symbol can be used by white supremacists, some of which covertly use the symbol to indicate their support for white nationalism \citep{bronsdonWhatDifferentTwitter}. 
Hence, emojis can indicate how these groups/movements interact with different personality types. 
We determine each emoji's frequency in a user's tweets and include these frequencies as predictors in upsampled LR models. 
Performing the same stepwise feature selection algorithm as above, we display the 12 most important predictors from the remaining models in Figure \ref{fig:VIP_plots_emoji}.

\begin{figure}[ht]
  \centering
  \begin{subfigure}{.49\linewidth}
    \centering
    \includegraphics[width=0.9\linewidth]{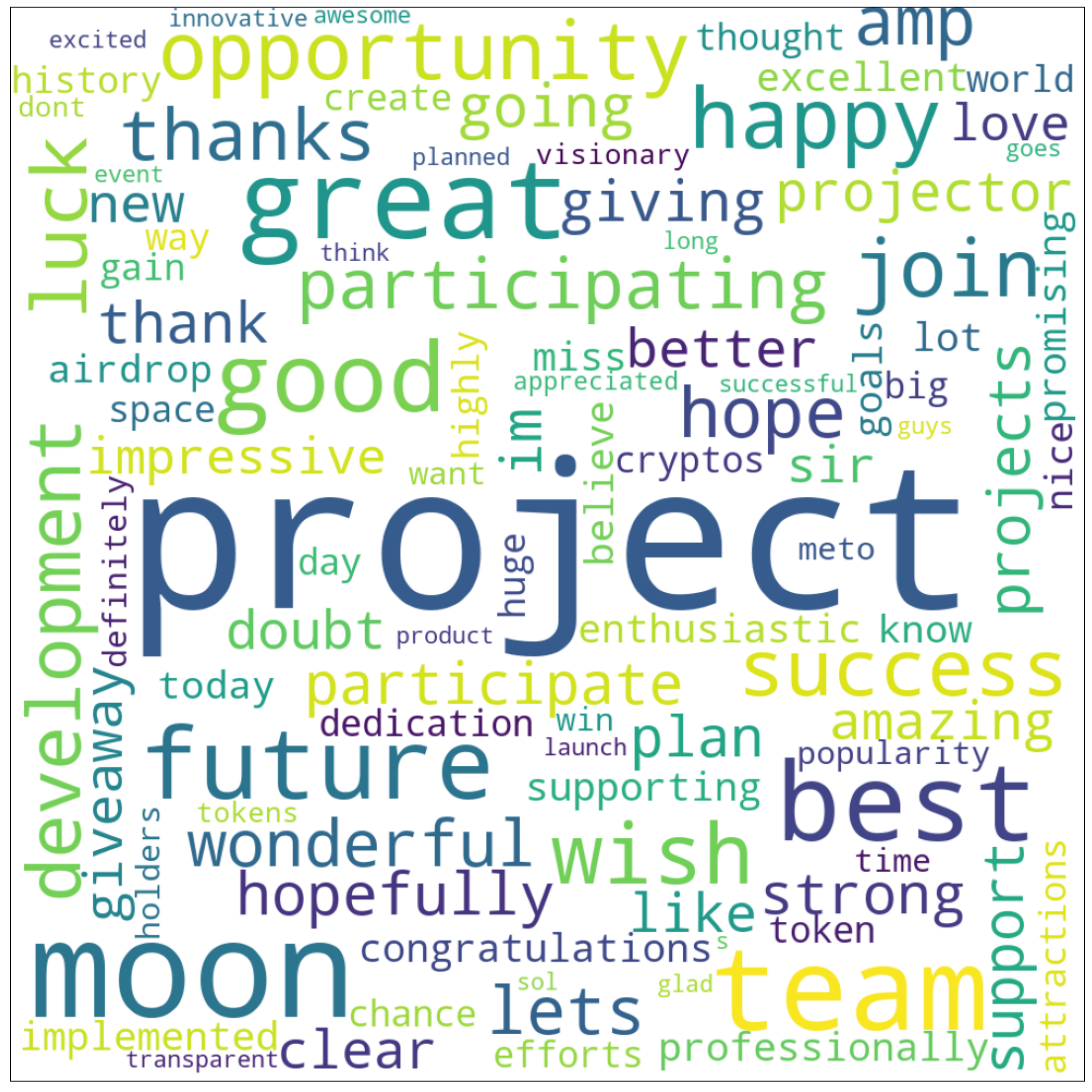}
    \caption{Rocket Ship Emoji}
    \label{fig:rocket_emoji_cloud}
  \end{subfigure}
  \begin{subfigure}{.49\linewidth}
    \centering
    \includegraphics[width=0.9\linewidth]{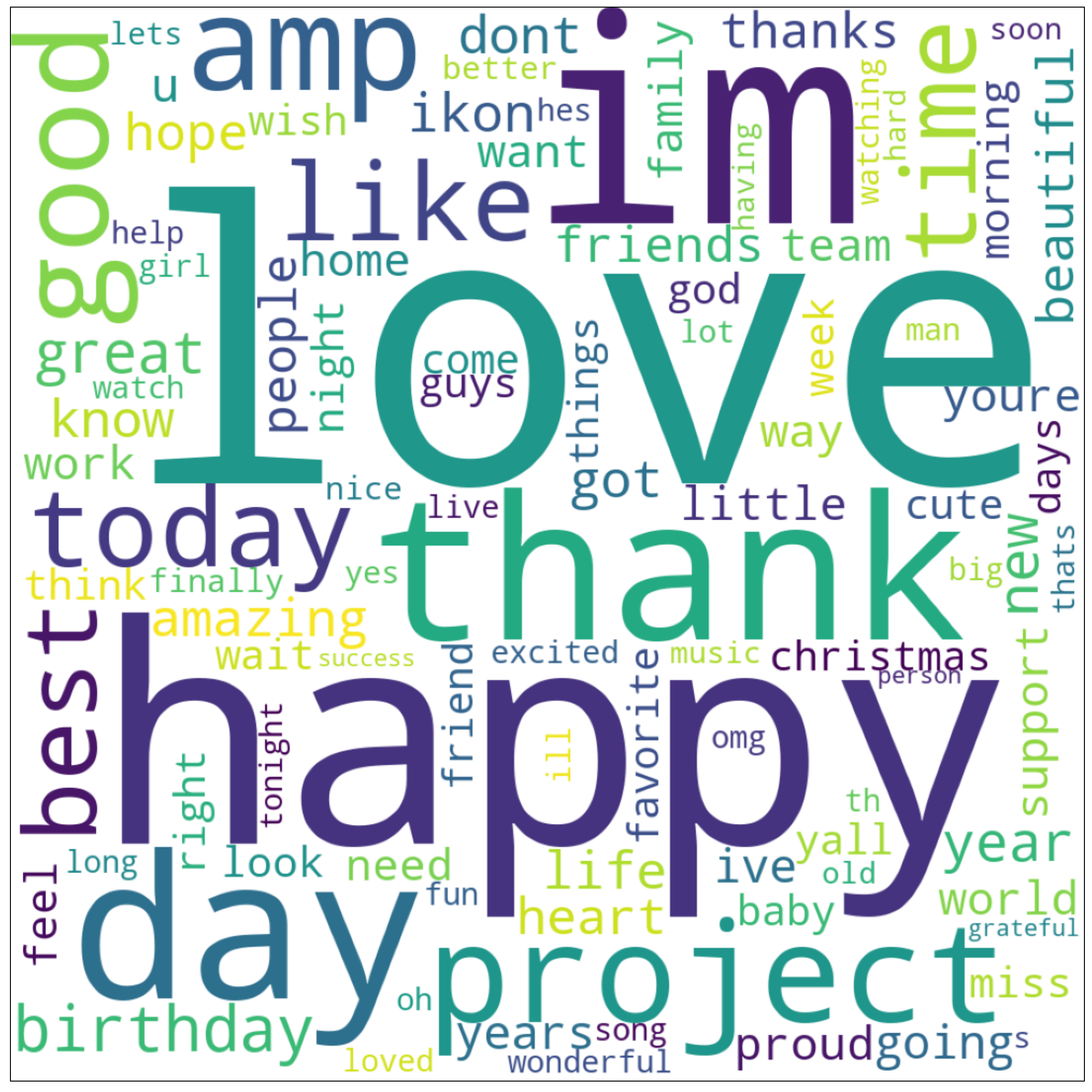}
    \caption{Red Heart Emoji}
    \label{fig:heart_emoji_cloud}
  \end{subfigure}
  \caption{Word clouds of tweets/quotes containing specific emojis in our dataset: rocket ship (left) and red heart (right). Note that we remove stopwords as they do not provide much context for the tweets.}
  \label{fig:emoji_word_clouds}
\end{figure}

The rocket ship emoji is one the top 12 most important predictors across all models. 
An increase in this emoji's usage implies a higher likelihood of an account being introverted, intuitive, feelings-orientated and perceiving. 
The rocket ship emoji has been used by finance enthusiasts who use the emoji to denote a fast increase in a particular stock or cryptocurrency. 
Hence, it is possible that we are observing crypto enthusiasts to be more introverted, intuitive, feelings-orientated and perceiving. 
However, this emoji has other meanings like as an actual rocket ship, so we explore created word clouds of tweets containing the rocket emoji (Figure \ref{fig:rocket_emoji_cloud}), as well as the red heart emoji (Figure \ref{fig:heart_emoji_cloud}). 
The rocket ship generally appears in crypto-related tweets discussing `projects', `great opportunities', `developments' and `cryptos'.
However, it also appears in tweets discussing the `moon' and `space'. 
The red heart emoji mainly appears in emotive tweets discussing `love' and `happiness'. 
A number of the emojis making an account more introverted are sad/upset emojis, whereas no sad/upset emojis make an account more extroverted. 
This further confirms Figure \ref{fig:EI_vip} which suggested that extroverts prefer to display positive emotion online.

Next we consider the importance of different feature groups (including the BERT features) and discuss whether different groups of features are more informative in our models. 
Again, we fit an upsampled LR model to all features and perform stepwise feature selection on each model.
We use the same thresholds to accept and remove features. 
We then measure the feature group importance using the number of remaining features in each feature group after selection. 
For each model, Table \ref{tab:remaining_features} displays number of predictors (in each feature group) and proportion that remain after stepwise feature selection. 
This proportion can be considered a measure of the importance of each feature group, which is not biased by the number of features in each group. 
We introduce a statistical framework to determine whether different groups of features are more informative for our data, by performing a Chi-Squared Test on the number of features retained and excluded from each model. We test the null hypothesis that each feature group is equally informative (per feature) and include the $p$-values from the Chi-Square Test in the captions of Table \ref{tab:remaining_features}.

\begin{table}[htb!]
\scriptsize
  \begin{subtable}{0.5\linewidth}
  \centering
  \begin{tabular}{c|c|c}
      \specialcell{Feature \\ Type} & \specialcell{\#} & \specialcell{Prop. \\ Retained} \\
      \hline
      SM & 4 & 36.4\% \\
      LIWC & 15 & 20.3\% \\
      BERT & 176 & 22.9\% \\
      Botometer & 1 & 14.3\% \\
      VADER & 2 & 33.3\% \\
      \hline
      Total & 198 & 22.9\% \\
  \end{tabular}
  \label{tab:EI_features}
  \caption{E/I $\br{p = 0.720}$}
  \end{subtable}%
  \begin{subtable}{0.5\linewidth}
    \centering
    \begin{tabular}{c|c|c}
        \specialcell{Feature \\ Type} & \specialcell{\#} & \specialcell{Prop. \\ Retained} \\
        \hline
        SM & 7 & 63.6\% \\
        LIWC & 18 & 24.3\% \\
        BERT & 217 & 28.3\% \\
        Botometer & 0 & 0.00\% \\
        VADER & 1 & 16.7\% \\
        \hline
        Total & 243 & 28.1\% \\
    \end{tabular}
    \label{tab:NS_features}
    \caption{N/S $\br{p = 0.032}$}
    \end{subtable}%
  \medskip

  \begin{subtable}{0.5\linewidth}
    \centering
    \begin{tabular}{c|c|c}
        \specialcell{Feature \\ Type} & \specialcell{\#} & \specialcell{Prop. \\ Retained} \\
        \hline
        SM & 5 & 45.5\% \\
        LIWC & 11 & 14.9\% \\
        BERT & 124 & 16.1\% \\
        Botometer & 1 & 14.3\% \\
        VADER & 3 & 50.0\% \\
        \hline
        Total & 144 & 16.6\% \\
    \end{tabular}
    \label{tab:TF_features}
    \caption{T/F $\br{p = 0.019}$}
    \end{subtable}%
    \begin{subtable}{0.5\linewidth}
      \centering
      \begin{tabular}{c|c|c}
          \specialcell{Feature \\ Type} & \specialcell{\#} & \specialcell{Prop. \\ Retained} \\
          \hline
          SM & 4 & 36.4\% \\
          LIWC & 8 & 10.8\% \\
          BERT & 112 & 14.6\% \\
          Botometer & 0 & 0.00\% \\
          VADER & 0 & 0.00\% \\
          \hline
          Total & 124 & 14.3\% \\
      \end{tabular}
      \label{tab:JP_features}
      \caption{J/P $\br{p = 0.120}$}
      \end{subtable}%
  \caption{Number of features and proportion retained in each group after stepwise feature selection. $p$-values are from Chi-Squared Tests on the null hypothesis that each feature group is equally informative per feature.}
  \label{tab:remaining_features}
\end{table}

\begin{figure}[ht]
  \begin{center}
    \includegraphics[width=\linewidth]{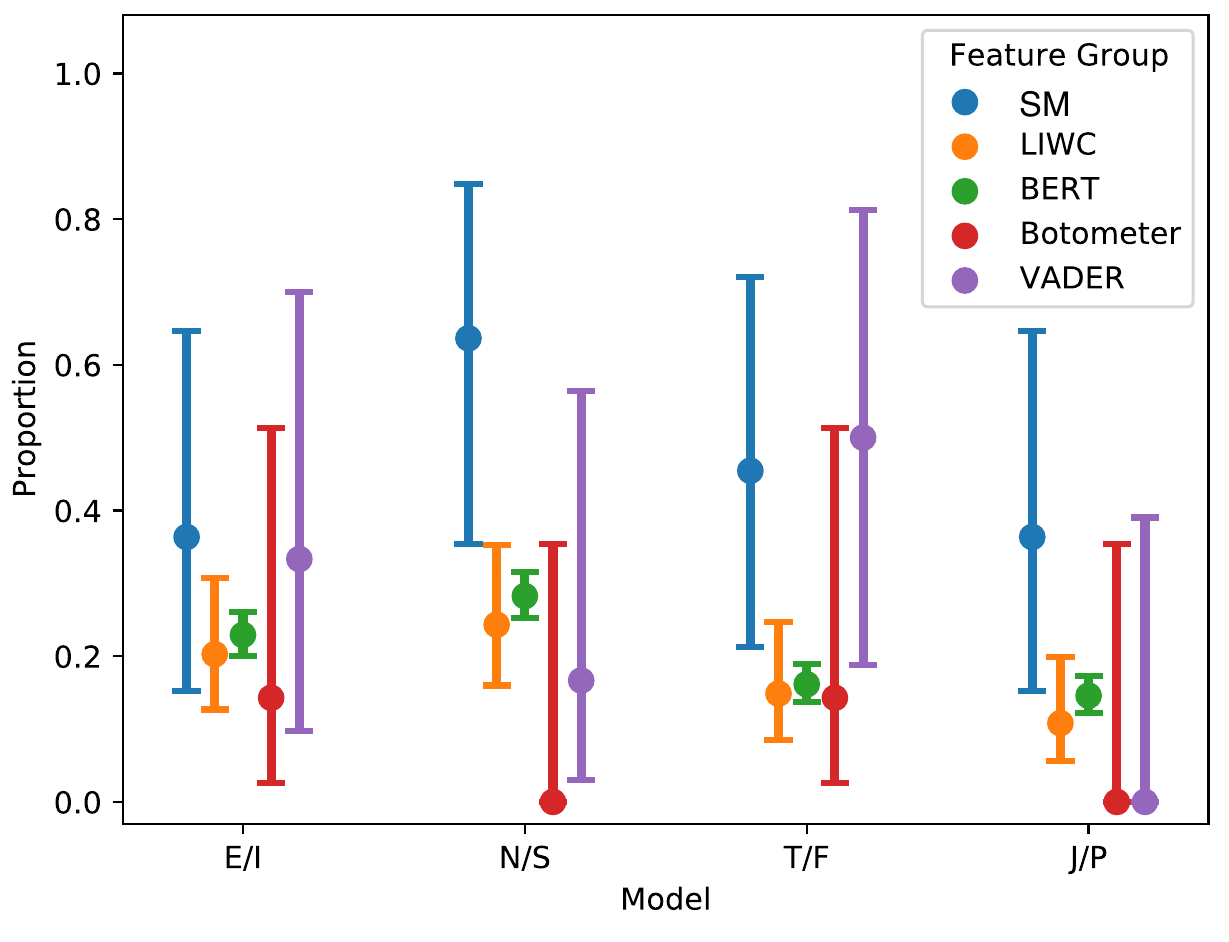}
  \end{center}
  \caption{95\% Wilson Score Binomial CIs for the proportion of retained features in each group. We use the Wilson Score version to correct for having zero successes in some cases.}
  \label{fig:feature_props}
\end{figure}

The number of features selected depends on the type of model. 
For instance, 243 features are selected in the N/S model, whereas only 124 features are selected in the J/P model. 
Interestingly, the N/S model is also the most accurate and the J/P model the least accurate, implying a positive relationship between accuracy and number of features retained. 
This is consistent with the remark that more features are retained in a model when they are more informative about the data. 
Moreover, the SM features are on average the most-retained across models. 
Conversely, the Botometer features have worst payoff across the four models, having the smallest proportion retained on average. 
The most interesting comparison is between the LIWC and BERT features, which both aim to describe linguistic properties about users. 
In each model, the BERT features are more highly retained. 
However, only the results from the N/S model and the T/F model are significant at the 5\% level. 
We therefore reject the null hypothesis that each feature group is equally as informative (per feature) for the N/S and T/F models. 
However, the Chi-Squared Test does not alone tell us what feature groups perform significantly better, so we perform individual confidence intervals (CIs) for the binomial proportions of accepting/rejecting features in each group using the Wilson Score interval \citep{reedBetterBinomialConfidence2007}. 
The CIs for each feature group and model are displayed in Figure \ref{fig:feature_props}.

For the I/S model, the 95\% CI for the SM features lies completely above those for LIWC and BERT. 
This indicates that SM features are more informative (per feature) than LIWC and BERT features at the 5\% level for this dichotomy. 
Attributes about a user's account are therefore sometimes more important than the language they use when modelling personality. 
This is also validated by the results for the T/F model, where the 95\% CI for the SM features and VADER features lie completely above the 95\% CI for the BERT features. 
We likely observe these results because the textual features are all fairly correlated with each other. Moreover, there is no evidence to suggest that BERT features are more informative than LIWC features in determining a Myers-Briggs personality type.

\section{Conclusion}
This paper contributes a labelled Twitter dataset of personality types and framework to model the personality types of these users. 
To our knowledge, this is the largest available Twitter dataset of labelled Myers-Briggs Personality Types.
Our data collection techniques avoid the long, cumbersome questionnaires used in other research. 
Additionally, we develop a statistical framework which combines NLP and mathematical models to model/predict users' personality type. 
While we applied this framework to personality types, it can model any labelled characteristics of online accounts -- political opinions, psychological properties or propensity to adopt an opinion. 
Using this framework, we analyse and compare a number of different models. 
Since personality types in our dataset are unbalanced, we compare different weighting/sampling techniques to deal with class imbalances. 
We discover that class imbalances are common in these types of datasets, yet are often overlooked. 
Because of this, we demonstrate why personality prediction models appear more accurate than they are, and demonstrate why digital footprints may be less informative of personality type than models suggest. 

\section*{Acknowledgments}
LM acknowledges support from the Australian Government through the Australian Research Council’s Discovery Projects funding scheme (project DP210103700).

\bibliography{main}

\end{document}